\documentclass[conference]{IEEEtran}

\usepackage[ruled]{algorithm2e}
\usepackage{stmaryrd}
\usepackage{graphicx}
\usepackage{color}
\usepackage{comment}
\usepackage{url}
\usepackage{wrapfig}
\usepackage{subfigure}
\usepackage{tabularx}
\usepackage{algpseudocode}
\usepackage{wrapfig}

\begin{document}

\title{The Case for High-Accuracy Classification:\\\textit{Think Small, Think Many!}}

\author{\IEEEauthorblockN{Mohammad Hosseini, Mahmudul Hasan}
\IEEEauthorblockA{
  \begin{tabular}{ccc}
    Comcast AI Labs, USA \\
  \end{tabular}
}}

\maketitle

\begin{abstract}
To facilitate implementation of high-accuracy deep neural networks especially on resource-constrained devices, maintaining low computation requirements is crucial. Using very deep models for classification purposes not only decreases the neural network training speed and increases the inference time, but also need more data for higher prediction accuracy and to mitigate false positives.

In this paper, we propose an efficient and lightweight deep classification ensemble structure based on a combination of simple color features, which is particularly designed for ``high-accuracy" image classifications with low false positives. We designed, implemented, and evaluated our approach for explosion detection use-case applied to images and videos. Our evaluation results based on a large test test show considerable improvements on the prediction accuracy compared to the popular ResNet-50 model, while benefiting from 7.64x faster inference and lower computation cost.

While we applied our approach to explosion detection, our approach is general and can be applied to other similar classification use cases as well. Given the insight gained from our experiments, we hence propose a \textit{"think small, think many"} philosophy in classification scenarios: that transforming a single, large, monolithic deep model into a verification-based step model ensemble of multiple small, simple, lightweight models with narrowed-down color spaces can possibly lead to predictions with higher accuracy.

\end{abstract}
\IEEEpeerreviewmaketitle

\section{Introduction}
\label{Introduction}


In the recent years, there has been a significant interest in applying deep neural networks to address many important real-world computer vision problems. As a result of seeking higher prediction accuracy, neural network complexity increased dramatically over time. Unfortunately, many times the resulting complex and larger neural networks only exacerbates the problem by consuming more memory and computational resources. Yet, larger datasets are needed to provide higher prediction accuracy and address false positives. While techniques like ensemble learning are designed to improve the classification accuracy, many of these models create a subset dataset from the original dataset, which is then used to make predictions on the whole dataset. Therefore, there is a high chance that these models will provide the same result since they are getting the same input. So in many cases, ensemble models may not perform well without feature engineering.

In this paper, we propose a novel ensemble structure consisting of two lightweight deep models based on combining simple color features, which are designed to perform high-accuracy and fast classification applied to both image and video classification use cases. Our design uses a verification-based combination of two lightweight deep models, each individual model making predictions on a different color feature: the main color-based model which operates based on 3 RGB color channels, and a secondary structure-oriented model which operates based on a single grayscale channel, focusing more on the shape of the object through intensity than learning about their dominant colors. We implemented and evaluated our approach for explosion detection use case where video scenes containing explosion are identified. Our evaluation results based on experiments on a large test set show considerable improvements on the classification accuracy over using a ResNet-50 model, while benefiting from the structural simplicity and the significant reduction in the inference time as well as the computation cost by a factor of 7.64.

\begin{figure}[!t]
\centering
\includegraphics[width=1\columnwidth]{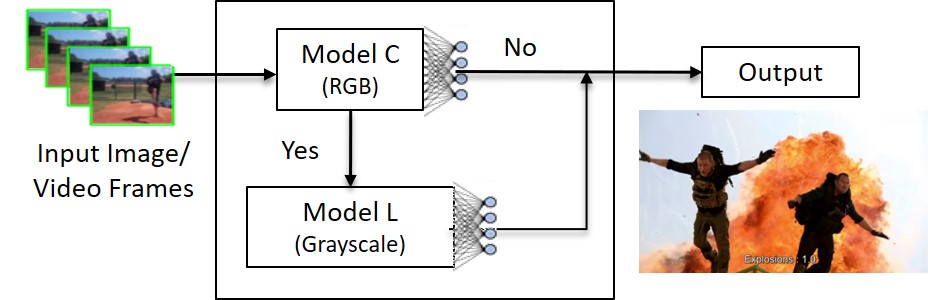}
\vspace{-0.5cm}
\caption{An abstract structure of our verification-based ensemble model.}

\label{fig:abstract}
\vspace{-0.5cm}
\end{figure}

\begin{figure}[!b]
\centering
\includegraphics[width=0.48\columnwidth]{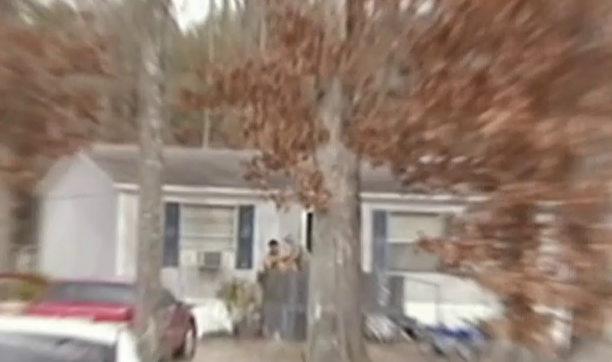}~\includegraphics[width=0.48\columnwidth]{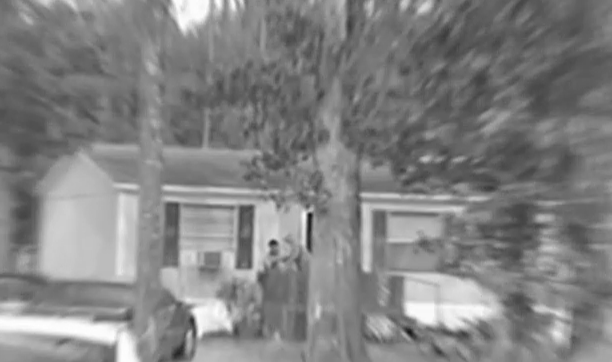}
\caption{An example false detection when making prediction using a grayscale model. When in grayscale, the image structure resembles an explosion.}
\vspace{-0.1cm}
\label{fig:falsepositive}
\end{figure}
While our approach is applied to explosion detection scenarios, we believe our approach can be generalized to other similar image and video classification use cases as well. Given the insights gained from our evaluations, we further make an argument to "\textit{think small, think many}", aimed to beat the complexity of large models with the simplicity of smaller ones- we express that replacing a single, large, monolithic deep model into a verification-based hierarchy of multiple simple, small, and lightweight models with step-wise contracting color spaces can possibly result in more accurate predictions.

\begin{figure*}[!t]
\centering
\vspace{-0.2cm}
\includegraphics[width=1\textwidth]{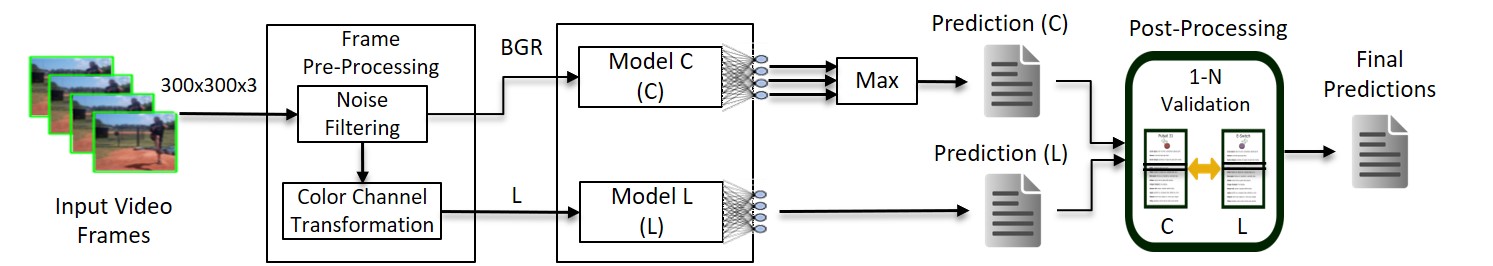}
\vspace{-0.5cm}
\caption{Details of our proposed ensemble extended towards video analytics. C and L represent colored and gray-scale features.}
\label{fig:details}
\vspace{-0.3cm}
\end{figure*}

\begin{figure*}[!t]
\centering
\includegraphics[width=0.9\textwidth]{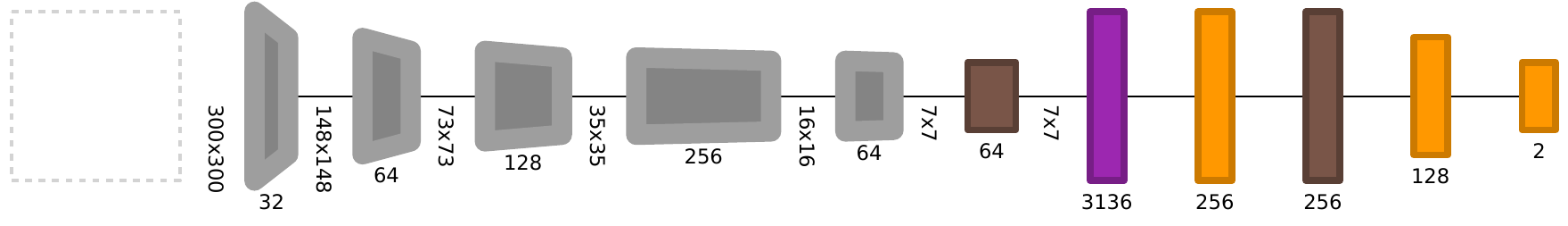}
\vspace{-0.3cm}
\includegraphics[width=0.9\textwidth]{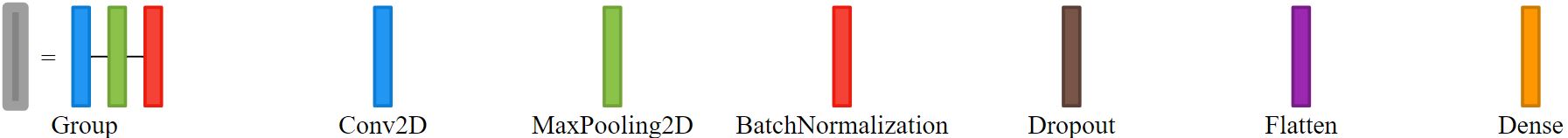}
\vspace{0.2cm}
\caption{The internal architecture of our base models. Model C and L both use the same architecture.}
\label{fig:model}
\vspace{-0.4cm}
\end{figure*}

The rest of the paper is organized as follows. In Section II, we briefly cover some background information and related work. Section III, we explain the design and the architecture of our ensemble approach. Section IV describes our experimental setup and evaluation results. In Section V, we discuss our argument, and illustrate how our proposal can be further generalized. We finally conclude the paper in Section VI.

\section{Related Work}
Ensemble models are generally aimed to improve the classification accuracy. Multiple models are used to make predictions for each data point, with each sub-model trained individually through variations in the input data. The predictions by each model are considered as a vote, where all votes can later be fused as a single, unified prediction and classification decision.

To achieve this, various techniques exist to combine predictions from multiple models. Voting, averaging, bagging and boosting are among the widely-used ensemble techniques \cite{ensemble1, ensemble2, ju2018relative}. In max-voting for instance, each individual base model makes a prediction and votes for each sample. Only the sample class with the highest votes is included in the final predictive class. In the averaging approaches, the average predictions from individual models are calculated for each sample. In bagging techniques, the variance of the prediction model is decreased by random sampling and generating additional data in the training phase. In boosting, subsets of the original dataset are used to train multiple models which are then combined together in a specific way to boost the prediction. Unlike bagging, here the subset is not generated randomly. While effective, a requirement in many of these broad approaches is to create a subset dataset from the original dataset, which is used to make predictions on the whole dataset. So there is a high chance that these models will give the same result since they are getting the same input.  

A body of existing work studied various ensemble neural network techniques aimed to improve prediction accuracy. In \cite{tao2019deep}, the authors propose an ensemble algorithm to address overfitting through looking at the paths between clusters existing in the hidden spaces of neural networks. The authors in \cite{wasay2018mothernets} present MotherNets, which enable training of large and diverse neural network aimed to reduce the number of epochs needed to train an ensemble. In \cite{bonab2017less}, the authors propose a geometric framework for a priori determining the ensemble size through studying the effect of ensemble size with majority voting and optimal weighted voting aggregation rules.

In this paper, we argue that transforming a single, large, monolithic deep model into an ensemble of multiple smaller models can potentially enable higher accuracy, while benefiting from reduced training costs and faster inference time. Our ensemble approach differs from the existing approaches as it uses the same dataset to train the individual models, and enables a mechanism through a verification-based combination of multiple models and use of simple color features with negligible transformation complexity. Our design further allows extension towards video frames through a low-complexity post-fusing mechanism as well. Our approach is independent, and can further be combined with other ensemble techniques proposed in the related work.

\section{Methodology}
The gist of our paper is a verification-based combination of two lightweight deep models with step-wise contracting color spaces: a "Model C", a color-oriented model which focuses on color features using 3 RGB color channels, and a "Model L", which is more structure or shape-oriented and performs verification of Model C predictions using gray-scale (L) features. Figure \ref{fig:abstract} illustrates the abstract design of our verification-based ensemble structure. Our simple ensemble structure suggests that if Model C predicts an input sample image as negative, the overall prediction would result as negative, and only if Model C predicted an input sample as positive, the prediction will be verified or validated with Model L. Our proposed structure is experimentally designed based on two-fold insights:

Firstly, for many use cases such as those seen in our explosion image classification use case, we experienced that while use of a single model operating on RGB-based color features in overall provides a higher accuracy, at the same time can wrongly classify scenes such as sunlight, lamps, or light-emitting sources as explosions. Similarly, while employing an individual model based on grayscale features can eliminate such color-induced false positives, it further introduces other false positives which are similar to explosion or fire in structural shape. Through our experimentation, we experienced that a grayscale model can potentially identify bushes or trees, clay grounds, clouds, or steam as fire or explosion simply due to lack of knowledge on the color data. This problem is exacerbated in video frames due to motion blur. So experimentally, we realized that a combination of both color and grayscale intensity (which focuses more on the structure) can provide a higher classification accuracy. Figure \ref{fig:falsepositive} depicts an example false positive identified as explosion when using only the grayscale model.

Secondly, limited features, in this case, limited number of color spaces through removing chrominance and keeping only luminance as the color feature, would generally lead to lower learnability simply due to lower features to train on. This means the model would incur higher recall and lower precision. Our evaluation showed that passing predictions from a supposedly higher-precision model to a model with higher-recall can potentially lead to filtering out false positives and therefore, increase overall prediction accuracy.

Figure \ref{fig:details} illustrates the structural details of an extended version of our ensemble design which is further applied to video frames. Video frames are captured and resized to 300x300, and pixels along with their 3 color channels are forwarded to a frame pre-processing phase. Our choice of 300x300 input dimension was determined through a trade-off between the computation cost and the prediction accuracy. Noise filtering using Anti-Aliasing technique is applied on every frame as a part of the pre-processing phase. Color channels are extracted from each frame, producing RGB color features (signified as C). Separately, a Color Channel Transformation module transforms the 3 RGB channels to grayscale-only feature (signified as L). The original RGB features are passed to Model C directly, while the gray-scale features are passed to Model L. Each model produces a binary prediction output signifying positive (i.e. explosion in our use case) or negative (i.e. non-explosion). 
Once predictions from individual models are made, a post-processing technique is applied on the prediction results from Model C (predictions C) and Model L (predictions L). During the post-processing step, a validation-based mechanism is employed where Model C's positive predictions are further \textit{verified} by predictions made by Model L. Only if Model L also made a positive prediction, the overall prediction would be positive.

As a part of the extension toward videos, an additional packing mechanism is applied to Model C predictions. This step is only applicable to videos where consecutive frames are available to further enhance the accuracy of the ensemble for video-based detection. However, our approach is general and does not require consecutive video frames. In the packing mechanism, the mode value of the prediction results of a pack of 3 frames on Model C are computed. The mode value represents the prediction which appears the most for each set of 3 frames. 
This value is then fused with the prediction from Model L through the same validation-based approach to derive the final outcome. To achieve that, an 1-N validation approach is employed. For every positive frame $i$ (labeled as "explosion") in predictions C results, we check the 3 neighbor frames $i-1$, $i$, and $i+1$ in predictions L. If at least one of these 3 frames were positive in prediction L results (labeled as "explosion"), the final prediction would be positive. While our video-specific post-processing approach can be generalized to other numbers of neighbor, such as 1 or 5, we found our choice of 3 to enable an efficient trade-off between the final precision and recall values.

Figure \ref{fig:model} depicts the internal architecture of each of the two models. Both Model C and Model L are feed forward convolutional neural network consisting of multiple groups of a 2D convolution layer (Conv2D), followed by a Max-Pooling layer and a Batch Normalization layer. There are five layers with the number of convolution filters and kernel sizes as shown in Figure \ref{fig:model}.  
We apply batch normalization to standardize the inputs passed to the next layer, which accelerates the training and provides regularization to reduce the generalization error. A dropout mechanism with a rate of 0.2 is also applied to help with preventing possible overfitting. Through a Faltten layer, output features are then flatten into a 1-dimensional array for entering the next layer. The produced data is then passed through three dense layers and another dropout layer to achieve the final binary prediction. We used Rectified Linear Unit (ReLU) as the activation function for all the Conv2D and Dense layers. While we designed this specific sequential neural network as our base model, we believe other lightweight models such as MobileNetV2 \cite{mobilenetv2}, SqueezeNet \cite{squeezenet}, or ShuffleNet \cite{shufflenet} can also be employed as the base model for our ensemble design as well.

\begin{figure}[!t]
\centering
\vspace{-0.3cm}
\includegraphics[trim=.9in 3.8in .9in 3.6in, width=0.9\columnwidth]{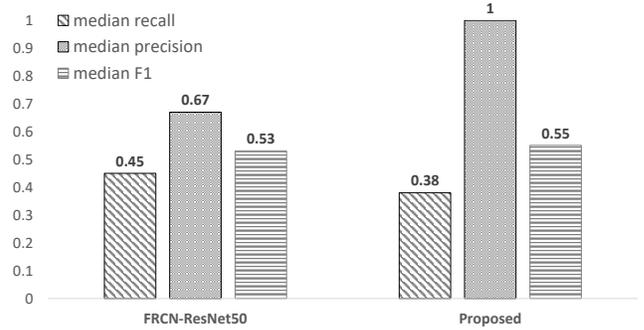}
\caption{Comparison of our proposed approach against the ResNet-50 model as used as the back-end of a Faster R-CNN detection network.}
\label{fig:evaluation}
\end{figure}

\begin{figure}[!t]
\centering
\includegraphics[trim=.75in 2.6in .75in 2.8in, width=.487\columnwidth]{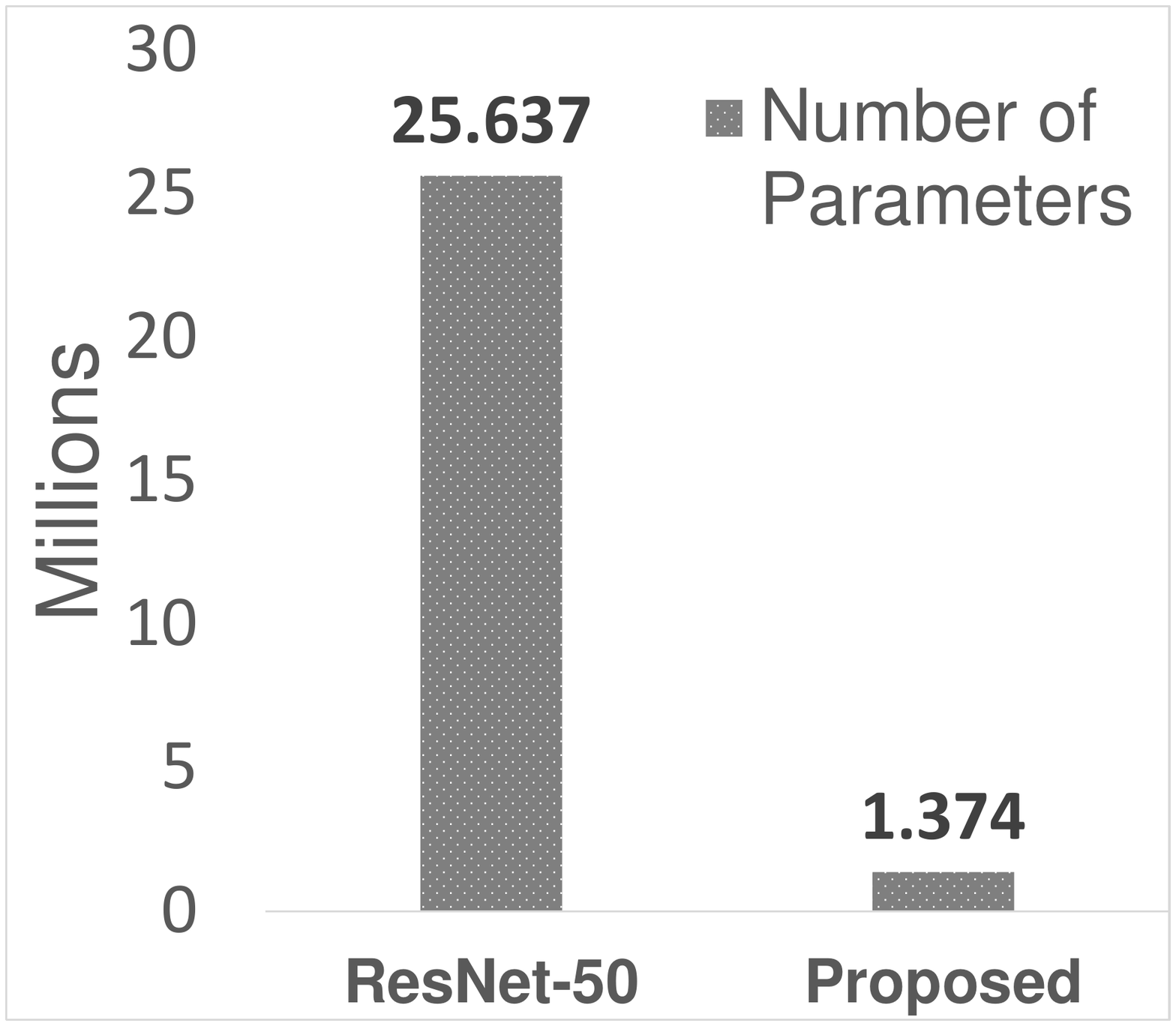}
\includegraphics[trim=.75in 2.6in .75in 2.8in, width=.49\columnwidth]{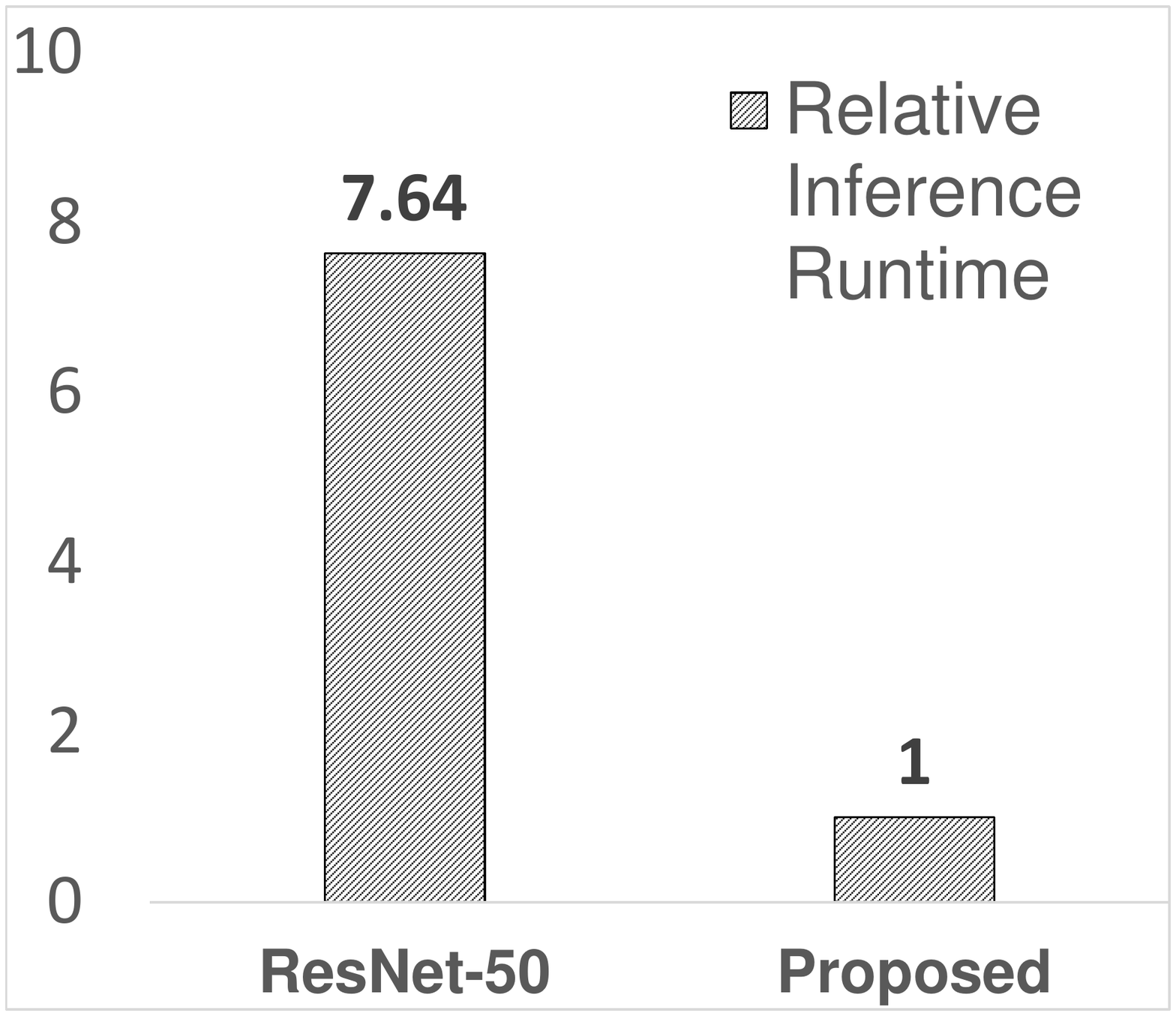}
\caption{The number of parameters (left) and the inference runtime (right) of our proposed ensemble relative to the popular ResNet-50 architecture.}
\vspace{-0.3cm}
\label{fig:paramsAndtime}
\end{figure}

\section{Evaluation}
As our goal was primarily to perform detection on videos, we created a large dataset of around 14,000 images, to train our models, consisting of around 8,000 negative and 6,000 positive images from real explosion footage originated from frames of videos. We split the dataset into training and validation sets, with the validation set consisting of 20\% of the whole dataset. We implemented the neural networks and the ensemble architecture on an Intel X86 64bit machine running Ubuntu 14.04.5 LTS, with Keras 2.3.1 using Tensorflow 1.13.1 back-end. We trained each of our models for 400 epochs, and saved the best model with lowest reported validation loss error at the end of the training phase. Evaluations of the proposed ensemble methods were conducted against the popular ResNet-50 architecture on a set of 15 test videos of various contexts provided by our vendor. The test videos include episodes of MacGyver, Britannia, NCIS: Los Angeles, and other popular TV series, encoded in 720p and 1080p resolutions, with an average duration of around 52 minutes for each video and average number of 78,750 frames per video. Human operators inspected the videos in multiple rounds to provide ground truth data with the time intervals of where explosion happened, where an average of 10.75 distinct explosion scenes were recorded as ground truth for an average test video.

As a part of our evaluation, we compared the accuracy results of our proposed ensemble model and the ResNet-50 model as used as the back-end of a state-of-the-art Faster R-CNN detection network \cite{frcn}. Figure \ref{fig:evaluation} compares the median precision, recall and F1 score metrics of the two models on the classification task. As the ground truth data provided by the human operators were recorded as time intervals, we converted the detection's frames to timestamps and considered a match if a detection was within a second of the recorded ground truth time.

\begin{figure}[!b]
\centering
\vspace{-0.5cm}
\includegraphics[width=.8\columnwidth]{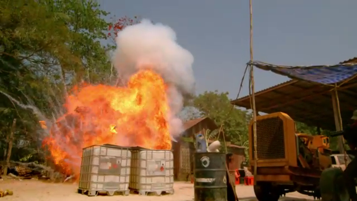}
\caption{An example of a correctly detected explosion scene on a test video.}
\vspace{-0.1cm}
\label{fig:example}
\end{figure}
Figure \ref{fig:paramsAndtime} shows how the number of parameters and inference time of our proposed ensemble method compares with the popular ResNet-50 architecture. On an average video, our proposed approach was able to achieve a 100\% precision which is significantly higher than the 67\% precision made by the popular ResNet-50 model as used as the back-end of a Faster R-CNN detection network, indicating the many false positives our approach was able to eliminate. Our high-accuracy detections can potentially save hundreds of hours of video-check operators by removing the need to verify false detections in the reference videos through manual inspections. In addition, our proposed structure is significantly lighter with almost 19x fewer parameters, with the ability to decrease inference run-time by a large factor, almost 7.64x faster compared to the complex ResNet-50 model. Figure \ref{fig:example} illustrates an example explosion scene correctly detected.

\begin{figure}[!t]
\centering
\vspace{-0.3cm}
\includegraphics[width=1\columnwidth]{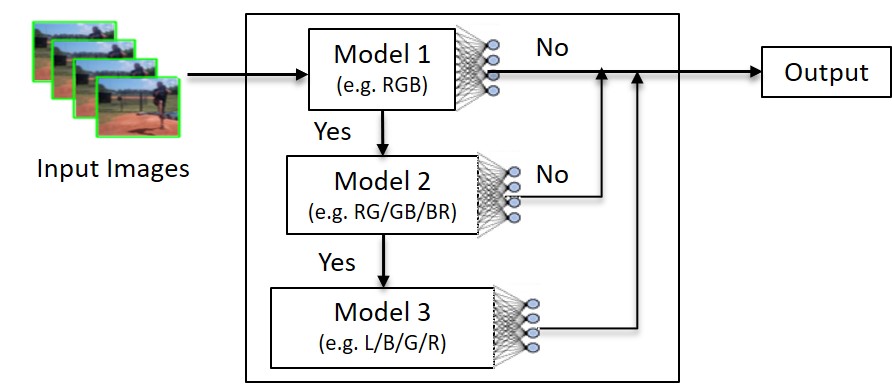}
\vspace{-0.5cm}
\caption{An abstract extension of our proposed ensemble design.}
\label{fig:extension}
\vspace{-0.5cm}
\end{figure}

\section{Discussion}
We believe that the design and experiments that we conducted on the explosion detection use case is general and can be applied to other similar image classification use cases as well. We therefore propose an open call and invite researchers to apply our design to other image or video classification tasks, especially those involving detection of non-rigid objects where color might be a dominant specification of the object, for instance blood gore detection, smoke detection, fire detection, steam detection, etc. Given the insights gained from our experiments, we hence propose a \textit{"think small, think many"} strategy in classification scenarios; we argue that transforming a single, large, monolithic deep model into a verification-based step-model ensemble of multiples of small, simple, and lightweight models with narrowed-down features, like our shrinking color spaces, can lead to predictions with higher accuracy. In this paper, we illustrate that our ensemble design was founded upon two base models, a primary and a secondary model, and used color spaces as the notion of features. The secondary model is fed with a limited set of the features delivered to the primary model, validating predictions made by the primary model. It is our belief that our design can be extended and generalized to a validation-based ensemble of 3 or more base models. Figure \ref{fig:extension} depicts a sample abstract illustration of our validation-based ensemble structure extended towards higher numbers of models. Model 2 validates predictions made by Model 1, and Model 3 further validates predictions made by Model 2, while the features passed (in our case, color spaces) get more limited as we iterate from Model 1 to Model 3. In this example extension, Model 1 operates on all three RGB channels, Model 2 operates on two color channels (RG, GB, or BR), and Model 3 performs validations only on a single color space, whether grayscale, or any of the R, G, or B channels. We believe this abstract concept can be generalized and extended to any feature set beyond only color spaces, which would be an avenue of exploration for the research community to consider.

\section{Conclusion}
\label{conclusion}
In this paper, we proposed an efficient and lightweight neural network ensemble structure, particularly designed for ``high-accuracy" detections in images and videos with low false positives. We designed, implemented, and evaluated our approach for an explosion detection use case. However, our approach is general and can be applied to other similar object classification use cases as well. Evaluations based on testing our models on a large test set show significant accuracy improvement over the popular ResNet-50 model, while benefiting from the simplicity, lower computation time, and faster inference time.

Given the insights gained from our experiments, we propose a \textit{"think small, think many"} philosophy when performing deep object classification tasks. We argue that transforming a single, large, monolithic deep model into a verification-based hierarchy of multiple small and lightweight models with contracting color channels can potentially lead to predictions with higher accuracy.

\bibliographystyle{IEEEtran}
\bibliography{sigproc}

\end{document}